\documentclass{article}

\PassOptionsToPackage{numbers, sort&compress}{natbib}

 \usepackage[dblblindworkshop, final]{neurips_2025}
\workshoptitle{Embodied World Models for Decision Making}

\usepackage[utf8]{inputenc} 
\usepackage[T1]{fontenc}    
\usepackage{hyperref}       
\usepackage{url}            
\usepackage{booktabs}       
\usepackage{amsfonts}       
\usepackage{nicefrac}       
\usepackage{microtype}      
\usepackage{xcolor}         
\usepackage{graphicx}
\usepackage{authblk}

\usepackage{titlesec}

\titlespacing{\paragraph}{0pt}{0pt}{0.5em}
\titlespacing{\section}{0pt}{0pt}{0pt}
\titlespacing{\subsection}{0pt}{0pt}{0pt}

\title{Opinion: Learning Intuitive Physics May Require More than Visual Data}

\author{%
   Ellen Su\thanks{Corresponding authors. Equal contribution.} \quad Solim LeGris $^*$ 
   \quad Todd M. Gureckis \quad Mengye Ren \\
   New York University \\
   \texttt{\{ellensu, solim.legris\}@nyu.edu}
}

\begin{document}

\maketitle

\vspace{-0.3in}
\begin{abstract}
\looseness=-1300
    Humans expertly navigate the world by building rich internal models founded on an intuitive understanding of physics. Meanwhile, despite training on vast quantities of internet video data, state-of-the-art deep learning models still fall short of human-level performance on intuitive physics benchmarks. This work investigates whether data distribution, rather than volume, is the key to learning these principles. We pretrain a Video Joint Embedding Predictive Architecture (V-JEPA) model on SAYCam, a developmentally realistic, egocentric video dataset partially capturing three children's everyday visual experiences. We find that training on this dataset, which represents 0.01\% of the data volume used to train SOTA models, does not lead to significant performance improvements on the IntPhys2 benchmark. Our results suggest that merely training on a developmentally realistic dataset is insufficient for current architectures to learn representations that support intuitive physics. We conclude that varying visual data volume and distribution alone may not be sufficient for building systems with artificial intuitive physics.

\end{abstract}

\section{Introduction}
A core aspect of human intelligence is the ability to engage flexibly with our environments \citep{pfeifer2006how, clark1997being, brooks1991intelligence}. This ability allows us to pursue goals, respond dynamically to changing stimuli, and navigate flexibly in new settings. Previous research in cognitive science suggests that this aspect of human intelligence emerges from the rich internal models people build of the surrounding world \citep{craik1943nature, johnson1983mental, lake2016buildingmachineslearnthink}. We revisit the claim that what makes our mental models \textit{useful} in the physical world is that they are constrained by internalized physical principles which reflect real-world dynamics \citep{battaglia2013simulation, ULLMAN2017649, gerstenberg2017intuitive, lake2016buildingmachineslearnthink}. As proposed by \citet{battaglia2013simulation}, humans may rely on an ``internal physics engine'': a cognitive mechanism that runs coarse, probabilistic simulations of how physical scenes evolve over short time periods. This mechanism possibly underpins our capacity to act and learn flexibly in novel environments. To achieve embodied AI systems with human-like abilities, we posit that the underlying world model should also demonstrate that it has internalized physical principles. However, state-of-the-art (SOTA) deep learning models still fail to approach human-level performance on intuitive physics benchmarks \citep{bordes2025intphys2benchmarkingintuitive}. Despite pretraining on millions of hours of video data and achieving impressive performance on motion understanding, video question-answering, and other downstream tasks \citep{assran2025vjepa2selfsupervisedvideo}, recent results by \citet{bordes2025intphys2benchmarkingintuitive} have provided evidence that these models perform just above chance at classifying between videos which are physically possible and impossible in the real world. 

For humans, decades of developmental psychology research have shown that physical reasoning is learned implicitly without any formal education and requires very little training data—infants under the age of one reliably demonstrate knowledge about object permanence and continuity through space and time \cite{Piaget1952Origins, piaget1954construction, baillargeon1991object, baillargeon2002acquisition}. Thus, our work mainly addresses an empirical question about data distribution: \textbf{to what extent can SOTA video models internalize physical principles by training on naturalistic and developmentally realistic data?} Rather than training video models on internet video and image datasets, we pretrain a V-JEPA model on SAYCam \citep{Sullivan2020SAYCamAL}, a developmentally realistic egocentric video dataset. SAYCam captures the richness and noisiness of early visual experiences, reflects a subset of the data that infants use to learn intuitive physics, and thus presents itself as a natural option for addressing learnability of physical principles with respect to data volume and distribution. We find that the V-JEPA architecture is unable to perform well on intuitive physics benchmarks even when trained on an approximation of the visual data distribution that human infants receive during development. Next, both SOTA models and our model, despite being trained on 0.01\% of the total data volume, achieve only slightly above chance performance on IntPhys2. In our analyses, we additionally show that intuitive physics tasks prove challenging for both the V-JEPA and VideoMAE learning algorithms. Our results suggest that visual datasets alone may be an insufficient training data source for current SOTA models to learn representations which support intuitive physics reasoning. In line with previous research, we propose that future work should utilize embodied datasets (egocentric, multimodal, action-annotated) and develop further innovations for model architectures. 

\section{Pretraining V-JEPA on a developmentally realistic dataset}
While much prior work has focused on domain-specific models for intuitive physics \citep{smithProbEOVnips2019, ULLMAN2017649}, recent architectural advancements have allowed for promising improvements to physical understanding in general-purpose deep learning models \citep{FinnGL16, garrido2025intuitivephysicsunderstandingemerges, assran2025vjepa2selfsupervisedvideo}. In particular, the Video Joint Embedding Predictive Architecture (V-JEPA) model \citep{Bardes2024Revisiting, assran2025vjepa2selfsupervisedvideo} has been hypothesized to learn flexible representations which may promote performance in physical reasoning tasks \citep{garrido2025intuitivephysicsunderstandingemerges}. The V-JEPA model is pretrained in a self-supervised manner and learns latent representations by predicting future states of masked frames. Recent work has demonstrated that it learns useful features which are applicable to many downstream tasks and domains \citep{Bardes2024Revisiting}. Thus, we opted for this model to explore our main question.

\subsection{SAYCam dataset}
We chose the SAYCam dataset as our training data as it contains 472 hours of egocentric videos collected from the point of view of children interacting in the real world. We hypothesized that SAYCam videos would provide richer learning signal for physical reasoning than internet videos from simulations or online collections since it more closely approximates the input that infants obtain, and sometimes actively seek, from the environment. Prior work has explored learnability of vision and language from  developmentally realistic and naturalistic data \cite{orhan2023learninghighlevelvisualrepresentations, orhan2024selfsupervisedlearningvideorepresentations}. In one instance, \citet{orhan2024selfsupervisedlearningvideorepresentations} demonstrated that generic deep learning models are able to learn meaningful representations of visual objects when trained on SAYCam in a self-supervised manner. Dataset details are provided in Appendix \ref{app:saycam}.

\subsection{Evaluating intuitive physics}
Intuitive physics understanding is difficult to measure. Without having direct access to the internal models of humans or AI systems, researchers often probe for specific knowledge or skills by analyzing behavioral patterns within a constrained and simplified environment~\cite{bear2021physion}. Prior work in developmental psychology has established the violation-of-expectation paradigm, which relies on infant looking times to determine which of a pair of visual events is more surprising, and thus less aligned with the expectations of the child \citep{margoni2024violation, BaillargeonSpelkeWasserman1985ObjectPermanence}. Following suit, many machine learning benchmarks evaluate video models on intuitive physics concepts by extracting a surprise metric from model outputs for each video frame \citep{bordes2025intphys2benchmarkingintuitive, riochet-intphys2022, weihs2022benchmarking, jassim2024graspnovelbenchmarkevaluating}. This metric assumes that, if the predictions of the video model capture real-world physical principles, then physically plausible videos should yield lower surprise scores than implausible videos. The computation of our metrics are detailed in Appendix \ref{app:surprise}. We chose to evaluate our model on the IntPhys2 benchmark, which was recently introduced by  \citet{bordes2025intphys2benchmarkingintuitive} and improves upon previous intuitive physics understanding benchmarks by incorporating dynamic shadows and lighting, natural occlusions, and both fixed and moving camera shots. Additional benchmark details are included in Appendix \ref{app:dataStats}. 

\subsection{Implementation details}
We pretrained a V-JEPA base model (215M parameters) with 16x16 patches at a spatial resolution of 224 (ViT-L/$16_{224}$) following the procedure set out in \citet{Bardes2024Revisiting} on the SAYCam dataset. Further details on model pretraining are described in Appendix \ref{app:implementation}. We compared our model against two other V-JEPA models with differently sized and distributed training datasets and against a Video-MAE model trained on SAYCam \citep{assran2025vjepa2selfsupervisedvideo,Bardes2024Revisiting, orhan2024selfsupervisedlearningvideorepresentations}. V-JEPA-2-H-VM22M (654M parameters)\footnote{\url{https://huggingface.co/facebook/vjepa2-vith-fpc64-256}} was pretrained on VideoMix22M \citep{assran2025vjepa2selfsupervisedvideo} (22 million samples), which comprises one egocentric video dataset \citep{goyal2017}, three exocentric video datasets \citep{kaykinetics2017, miech19howto100m, zellers2022merlot}, and 1 image dataset \citep{deng2009imagenet} and amounts to over 1.73 million hours of video data. Next, we evaluate V-JEPA-L-0-1-HowTo100M (326M parameters) which is pretrained by \citet{garrido2025intuitivephysicsunderstandingemerges} on a 0.1\% subsample of the HowTo100M dataset\citep{miech19howto100m}, and receives 128 hours of unique video data. While similarly sized to SAYCam, HowTo100M is made up of exocentric and general purpose online tutorial videos, allowing us to isolate the effect of video data distribution on task performance. Finally, the VideoMAE model released by \citet{orhan2024selfsupervisedlearningvideorepresentations} is trained in a self-supervised manner on the SAYCam dataset. We include the performance of this model to again isolate the effect of model architecture on performance. Our training code is available at \url{https://github.com/eysu35/vjepa} and our pretrained model is available for download at \url{https://huggingface.co/eys8549/vjepa-saycam}.
\begin{figure}[t]
\vspace{-0.3in}
\centering\includegraphics[width=\linewidth]{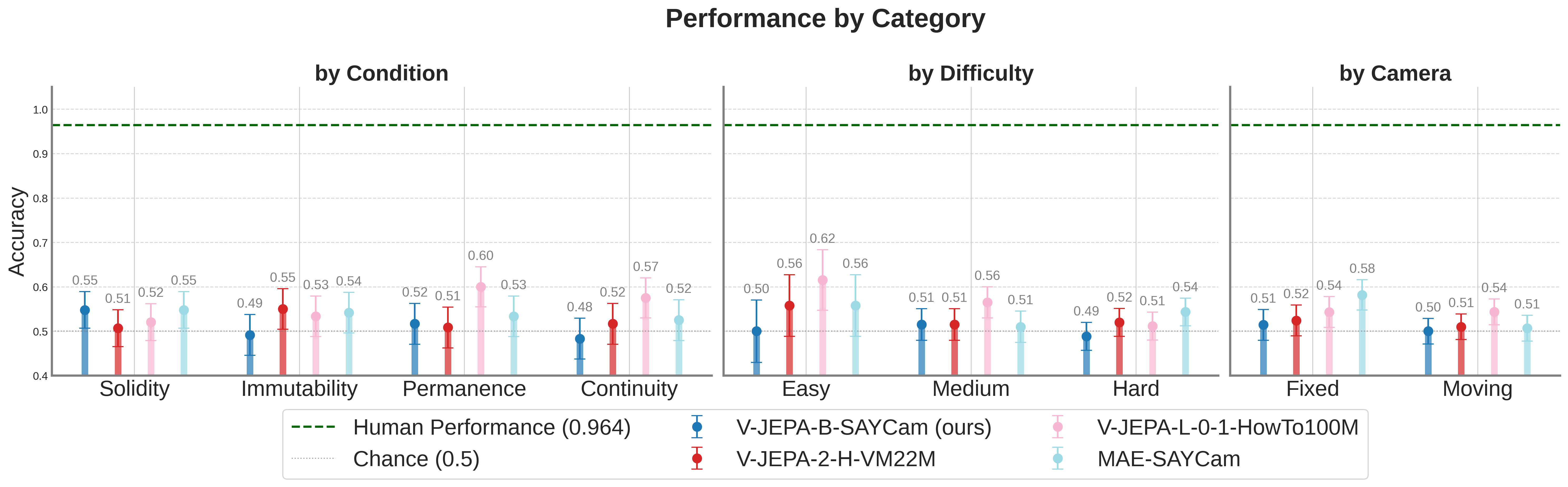}
    \caption{Breakdown of model accuracies on the IntPhys2 benchmark by physical condition, difficulty, and camera set up. All models achieve around chance performance at classifying between possible and impossible physical events with small variations in rank. Error bars mark standard error of the mean accuracy.}
    \label{fig:main-results}
\end{figure}
\section{V-JEPA pretrained on SAYCam yields similar performance to SOTA models}

We find that our model achieves similar performance to all other benchmarked models. All models are performing at or slightly above chance (V-JEPA-B-SAYCam (ours): 0.50, V-JEPA-2-H-VM22M: 0.52, MAE-SAYCam: 0.53, V-JEPA-L-0-1-HowTo100M: 0.54), suggesting that neither current large-scale internet video datasets nor human-scale naturalistic developmental video datasets are sufficient for V-JEPA architectures to learn intuitive physics concepts. The video pairs in the IntPhys2 dataset vary by physical concept, difficulty, and camera set up (see Figure~\ref{fig:main-results}), which allows for more fine-grained analysis compared to previous benchmarks. Despite sharing similar architectures and being trained on drastically different datasets, the models show consistently poor overall performance. In contrast, \citet{bordes2025intphys2benchmarkingintuitive} evaluated human participants and reported a near-perfect score of 96.44\% for overall classification accuracy. Our model, the VideoMAE model, and the V-JEPA large model were fully trained on a few hundred hours of unique video data. Given that this makes up less than 0.01\% of the data volume which V-JEPA-2-H-VM22M was trained on, we observe that the performance gain across intuitive physics tasks does not scale well with the size of training dataset. While pretraining on larger and more diverse video datasets may lead to richer learned representations and performance improvements in novel downstream tasks, our results highlight the fact that intuitive physics reasoning in video models remains a challenge currently unsolved by the volume of visual data alone. Comparison to V-JEPA-L-0-1-HowTo100M and MAE-SAYCam also indicate that data distribution and model architecture did not result in significant differences in task performance. 

\subsection{Surprise analysis}

\begin{figure}
\vspace{-0.3in}
    \centering
    \includegraphics[width=0.8\linewidth]{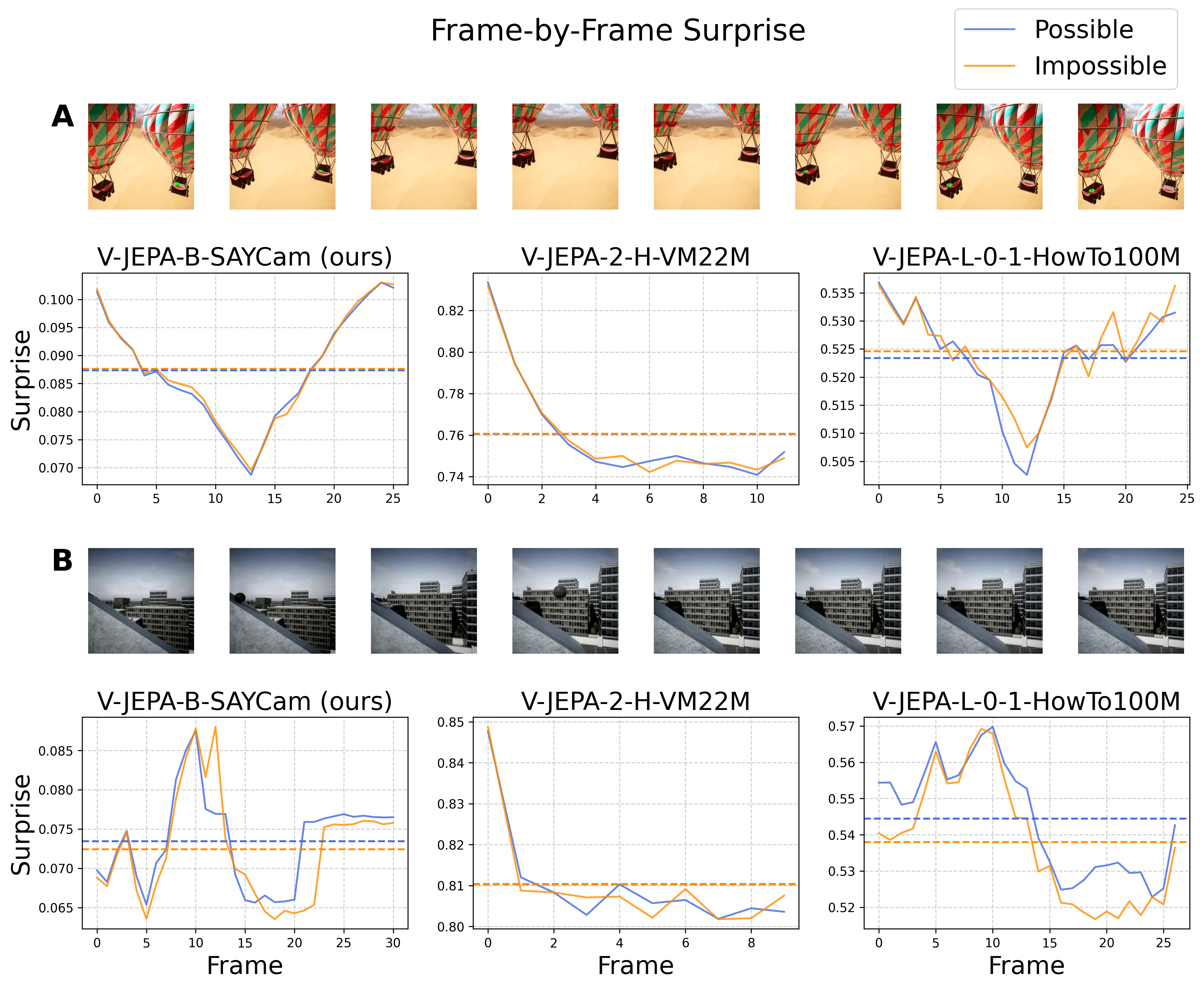}
    \caption{Model surprises for all V-JEPA models. \textbf{A)} Top: a subsample of video frames for an impossible video (violates object permanence). Bottom: Frame-by-frame surprise predictions for the corresponding possible/impossible video pair which all models classified correctly (average surprise for possible < average surprise for impossible). \textbf{B)} Top: a subsample of video frames for an impossible video (violates continuity). Bottom: Frame-by-frame surprise predictions for the corresponding possible/impossible video pair which all models classified incorrectly (average surprise for possible > average surprise for impossible). The surprise values were taken from the context which reported the best overall accuracy score. Dotted lines represent average surprise scores.}
    \label{fig:surprise}
\end{figure}

We conducted a qualitative analysis to compare the fine-grained behavioral patterns of surprise values between the V-JEPA models following the analysis in \citet{bordes2025intphys2benchmarkingintuitive}. The surprise patterns across video conditions shed light into whether the surprise metric captures higher-level concepts in the videos and whether the model classification is grounded in its understanding of these events. Figure \ref{fig:surprise} shows the predicted values over frames of a pair of correctly classified videos (panel A) and incorrectly classified videos (panel B). We observe that the surprise predictions from the two models trained on small-scale datasets follow similar patterns and respond to video contents more so than the model trained on VideoMix22M. Finally, for both video pairs, the surprise patterns outputted by all models are consistently similar for the possible and impossible videos, indicating that computing classification accuracy based on average surprise is highly sensitive to noise. 

\section{Discussion}

Although previous results suggested that human-level intuitive physics emerges from large-scale video datasets using latent representation learning \citep{garrido2025intuitivephysicsunderstandingemerges}, recently published benchmarks suggest that physical reasoning in world models remains a machine learning challenge \citep{bordes2025intphys2benchmarkingintuitive}. In this work, we explored whether the V-JEPA model could learn representations which reflect intuitive physics principles when trained on a video dataset that mimics the data a child is exposed to. Although we found that the model was able to match performance on some dimensions of the benchmark with significantly less data volume, our conclusion is that all evaluated models, which vary in size, pretraining data distribution, and architecture, are yet unable to learn representations which support intuitive physics reasoning. One limitation of our work is that the SAYCam dataset is of a drastically smaller size than the visual data children have access to and does not fully capture the range of visual experiences needed to learn intuitive physics. Thus, perhaps models would perform better on intuitive physics benchmarks if trained on a larger dataset of this distribution. Second, SAYCam only captures visual information and does not incorporate other aspects of embodiment which may be critical to learning intuitive physics. For example, perhaps children learn to understand physics by reconciling their self-directed movements with their visual fields, and thus incorporating motion or action data alongside visual data may be key to learning good representations for intuitive physics. Other datasets such as BabyView \citep{hsu2024babyview} might be more conducive to learning intuitive physics since they annotate egocentric videos with accelerometer data. In general, while egocentric and developmentally realistic video data are components of embodiment, we hypothesize that the signal captured in an agent's actions and motion would contribute meaningfully to internalizing physical principles in machine learning world models. The effort toward artificial intuitive physics aligns with the development of effective and efficient embodied AI systems that can operate freely in the world.

\section*{Acknowledgements}

The authors gratefully acknowledge helpful contributions and input from Quentin Garrido and Brenden Lake. This work was supported in part through the NYU IT High Performance Computing resources, services, and staff expertise.

\bibliography{references}
\bibliographystyle{abbrvnat}


\newpage
\appendix
\onecolumn
\section{SAYCam dataset} \label{app:saycam}
SAYCam is a longitudinal audiovisual dataset of head cam recordings collected from three young children (nicknamed S, A, and Y) from the ages of 6 to 32 months \citep{Sullivan2020SAYCamAL}. Each infant wore a head cam for approximately 2 hours per week over the course of approximately 2.5 years. In total, SAYCam contains 472 hours of video, with 194, 141, and 137 hours of video from S, A and Y, respectively. The data for each child consist of a set of continuous, natural and uninstructed recordings, each between 1 to 2 hours. All of the videos take place in the home environments of the children and include activities such as crawling, being held, lying, or sitting \cite{Sullivan2020SAYCamAL}. 

\section{Violation of Expectation Metric} \label{app:surprise}
The violation-of-expectation paradigm has been extensively used in developmental psychology to evaluate psychological constructs in preverbal infants \citep{BaillargeonSpelkeWasserman1985ObjectPermanence, margoni2024violation}. In a typical experimental setup, infants are presented with two similar visual scenes following habituation, where one of the scenes contains an event which is hypothesized to appear surprising only if the infant has some understanding of the underlying property being tested. Surprise is usually obtained by measuring relative looking time \citep{Spelke1985Preferential}, and is used to determine whether a concept violation has occurred. 

In recent machine learning literature, this notion of surprise has been adapted to serve as a proxy for model understanding \citep{garrido2025intuitivephysicsunderstandingemerges}. Both types of models we evaluated (pixel and latent prediction methods) can be evaluated in the same way, with the only difference being how the target of the prediction is encoded. For V-JEPA, we use the latent representations of the future obtained by encoding the video and then only keeping the future frames, while for the VideoMAE trained on SAYCam, the target is simply the normalized future of the video. Considering a video $V$ with frames $1, ..., T$ , a context encoder $f_{\theta}$ handling $C$ frames, a target encoder $g_{\psi}$ producing the ground truth $M$ future frames from the video, and a predictor predicting $M$ frames in the future, we can measure surprise at time $t$ as

\begin{equation}
    S_t = ||p_{\phi}(f_{\theta}(V_{t:t+C})) - g_{\psi}(V_{t:t+C+M})||_1
\end{equation}

Next, prior work reports different design choices for computing model accuracies from surprise values. Building on this work, we define our accuracy as the fraction of video pairs in which the average surprise (across all video frames) for a possible video is lower than the impossible video. As the evaluation is conducted over multiple context lengths as a hyperparameter, our final performance score is taken from the context which maximized the model accuracy for each condition following \citet{bordes2025intphys2benchmarkingintuitive} and \citet{garrido2025intuitivephysicsunderstandingemerges}. The average surprise per context is computed as,

\begin{equation}
    \texttt{AvgSurprise} = \frac{1}{T} \sum_{t\in \{ 1, 1+s, ..., T-(C+M)\}} S_t
\end{equation}

where $s$ is a stride parameter for the sliding window, reducing the amount of compute used to evaluate each video clip. For all our evaluations, we use $s=2$. 

\section{Benchmark details} \label{app:dataStats}

The IntPhys2 benchmark \citep{bordes2025intphys2benchmarkingintuitive} builds off of the first version by including pixel-to-pixel aligned pairs of videos simulating physically impossible and possible events. The creators of this benchmark generated all video pairs with a photorealistic simulation engine which avoids challenges like data leakage and the appearance of spurious features from video stitching. 

The benchmark contains contains three main splits: the ``Debug'' (30 pairs), ``Main'' (506 pairs) and ``Held-out'' (172 pairs) sets. Metadata containing labels, scene and difficulty among other attributes were only released for the debug and main splits. This new collection of generated videos are photorealistic and include dynamic shadows and lighting, natural occlusions, and both fixed and moving camera shots. All these components add richness to the dataset and establishes the benchmark as a more comprehensive and realistic evaluation benchmark for intuitive physics understanding. Following Bordes et al. \citep{bordes2025intphys2benchmarkingintuitive}, we evaluate our model on the ``Main'' data split which contains metadata annotations. Model performance is measured across object permanence, object immutability, spatio-temporal continuity, and solidity.

\section{Implementation Details}\label{app:implementation}
The pretraining procedure we followed was heavily influenced by the code and configurations set up in \citet{Bardes2024Revisiting} and \citet{garrido2025intuitivephysicsunderstandingemerges}. As we were comparing our pretrained model to the V-JEPA version 1 model trained on a large-scale video dataset, we mimicked the original training process as closely as possible. However, the drastic reduction in dataset size required us to make a few changes: reducing the training time from 200 (60,000 steps) to 40 epochs (12,000 steps), reducing the warm up period from 40 to 10 epochs, and fixing the weight decay at 0.04 rather. In addition, we partitioned the data from each child in SAYCam 80/20 into a train and validation set, monitored the loss curves of both to ensure convergence, and implemented early stopping to mitigate overfitting. Finally, we trained our model across 2 NVIDIA A100-SXM4-40GB GPUs. 
\end{document}